\documentclass[12pt]{elsarticle}

\usepackage{times,slashbox}
\usepackage{epsfig}
\usepackage{graphicx,xcolor}
\usepackage{amsfonts,amsmath,amssymb,amsthm,bbm}
\usepackage{enumitem}
\usepackage{booktabs}  
\usepackage{algorithm}
\usepackage{algpseudocode}

\usepackage[utf8]{inputenc} 
\usepackage[T1]{fontenc}    
\usepackage{url}            
\usepackage{booktabs}       
\usepackage{nicefrac}       
\usepackage{microtype}      

\newcommand{\Ucal}{\mathcal{U}}
\newcommand{\Acal}{\mathcal{A}}
\newcommand{\Gcal}{\mathcal{G}}
\newcommand{\Lcal}{\mathcal{L}}
\newcommand{\Pcal}{\mathcal{P}}

\newcommand{\R}{\mathbb{R}}
\newcommand{\C}{\mathbb{C}}
\newcommand{\E}{\mathbb{E}}

\newtheorem{theorem}{Theorem}

\newtheorem{proposition}[theorem]{Proposition}

\usepackage{hyperref}
\journal{Arxiv}
\begin{document}

\begin{frontmatter}

\title{Regime-Aware Time Weighting for Physics-Informed Neural Networks}

\author{Gabriel Turinici}
\address{Université Paris Dauphine - PSL\\
CEREMADE, \\ Place du Marechal de Lattre de Tassigny, Paris 75016, FRANCE
\\ \  \\ \today}
\ead[url]{https://turinici.com}
\ead{Gabriel.Turinici@dauphine.fr}

\begin{abstract}
We introduce a novel method to handle the time dimension when 
Physics-Informed Neural Networks (PINNs) 
are used to solve time-dependent differential equations; our proposal focuses on how time sampling and weighting strategies affect solution quality. While previous methods proposed heuristic time-weighting schemes, our approach is grounded in theoretical insights derived from the Lyapunov exponents, which quantify the sensitivity of solutions to perturbations over time. This principled methodology automatically adjusts weights based on the stability regime of the system - whether chaotic, periodic, or stable. Numerical experiments on challenging benchmarks, including the chaotic Lorenz system and the Burgers' equation, demonstrate the effectiveness and robustness of the proposed method. 
 Compared to existing techniques, our approach offers improved convergence and accuracy without requiring additional hyperparameter tuning. The findings underline the importance of incorporating causality and dynamical system behavior into PINN training strategies, providing a robust framework for solving time-dependent problems with enhanced reliability.

\end{abstract}

\end{frontmatter}

\section{Time in PINNs : introduction and previous works}

Physics-informed Neural Networks introduced in \cite{RAISSI2019686} proved to be a very successful paradigm invoked to solve complex mathematical equations such as time evolutions 
\cite{pinn_siam21,bae_option_2024_pinn,pinn_vol_surface24} \cite{mertikopoulos_almost_2020} possibly in high dimensions \cite{hu2024tackling}, 
 partial differential equations \cite{bae_option_2024_pinn}
 or control systems \cite{cuomo_scientific_2022,demo_extended_2023,mfg_pinn24}; this framework exploits the power of neural networks (including automatic differentiation) to represent the main unknown function which is the solution of some equation involving its derivatives. The PINNs can also incorporate different other elements such as (possibly noisy) measurements on the system or further data such as controls.

To improve PINN performance several leads have been followed; 
 Dolean et al. \cite{dolean_multilevel_2024} advocate multi-level domain decomposition techniques, 
A. Beguinet et al. \cite{beguinet_adrien_deep_2023} investigate how performance behaves for low regularity solutions, 
Sharma and Shankar
 \cite{sharma_accelerated_2022_pinn} studied meshless discretization,
Ibrahim et al. \cite{Ibrahim_2023} combine it with the "parareal" method~\cite{maday_parareal_2002},
  Cho et al. \cite{cho_separable_2023} consider separable network architecture while Wang et al. \cite{wang_neurips22_L2loss} analyze 
best norm to express the loss. We borrow here from all these perspectives but we focus on time-dependent problems and more specifically on how the sampling of time points (or their weighting in the loss) impacts the overall performance. 
Although special sampling methods such as the cubature \cite{lyons_cubature_2004} and quantization \cite{turinici_huber_energy_2024} could be used to take into account the 
particular characteristics of the time dimension, is has been long recognized that it  deserve special attention. In a highly influential work on 
Allen-Cahn and Cahn-Hilliard  
phase field equations \cite{adaptive_pinn20} Wight and Zhao introduce various sampling strategies in both space and time while McClenny and Braga-Neto
\cite{self_adaptive_pinn2023} make weights trainable which requires additional computational time; in the latter work the influence of the weights on the training loss is mediated by a function that becomes a hyper-parameter to be specified by the user.
In a related approach, 
\cite{pinn_causality} Wang et al. remarked that, especially for chaotic or near chaotic systems, the causality of the time dimension has to be respected and enforced to obtain good results. In particular they proposed that earlier times be given more weight in the loss functional, proportional to the exponential of the cumulative sum of the accumulated errors (up to that point in time, see formula~\eqref{eq:causal_weights_formula} below); a hyper-parameter denoted $\epsilon$ remains to be chosen and in practice it is iterated in a prescribed list. The procedure also requires some special termination criterion.  
On the other hand 
\cite{penwarden_unified_2023} Penwarden et al. also studied the time dimension and propose a scalable framework for causal sweeping strategies in PINNs. This and many other contribution recognized the necessity to combine PINNs sequentially in order to solve a problem set on a large time interval. While we agree with this perspective, we search here only for the best time weighting scheme on each of these individual intervals ; as such, our procedure can be combined with any other time sweeping protocol. Moreover, as the previous protocols have been rather prescriptive and proposed ad-hoc choices of time weighting, we focus here on the design of principled approaches to orient the choice of  weights.  Our procedure is on one hand flexible enough to recognize the main regime the evolution takes place into (stable, chaotic, periodic) and on the other hand it does not require additional hyper-parameters to adjust, which was a more laborious part of the previous approaches.

Our procedure works by using time weights as a distribution with density   proportional to 
$e^{- \int_0^t \lambda(s) ds}$; the quantity $\lambda(t)$ is related to a crude approximation of the Lyapunov exponent of the evolution equation (see Equation~\eqref{eq:estimator_ls_ref_first} for the detailed formula) and is obtained numerically, on the fly, during the PINN iterations. This choice allows to avoid precise estimation of the Lyapunov exponent which is computationally intensive and prone to errors.

Compared with other methods, our approach performs best on systems 
where the dynamics is non-linear and in particular where capturing stability or long-term behavior (e.g., via Lyapunov exponents) is essential. For some standard classes of differential equations the method might still work but may not offer a significant advantage over simpler or more traditional approaches like standard PINNs. However, since it has only a slight computational overhead using it will not affect significantly the overall efficiency of the numerical method. So, if there is no a priori information on the system's underlying dynamics we recommend to use the method.

The balance of the paper is the following: in section \ref{sec:framework} we introduce the general framework and in section \ref{sec:theory} we give the main theoretical insights that motivate our procedure. The numerical experiments are the object of section~\ref{sec:numerical}, followed in sections~\ref{sec:limitations} and~\ref{sec:conclusion} by concluding remarks.

\section{Presentation of the framework} \label{sec:framework}
\begin{figure}
\centering
\includegraphics[width=0.8\linewidth]{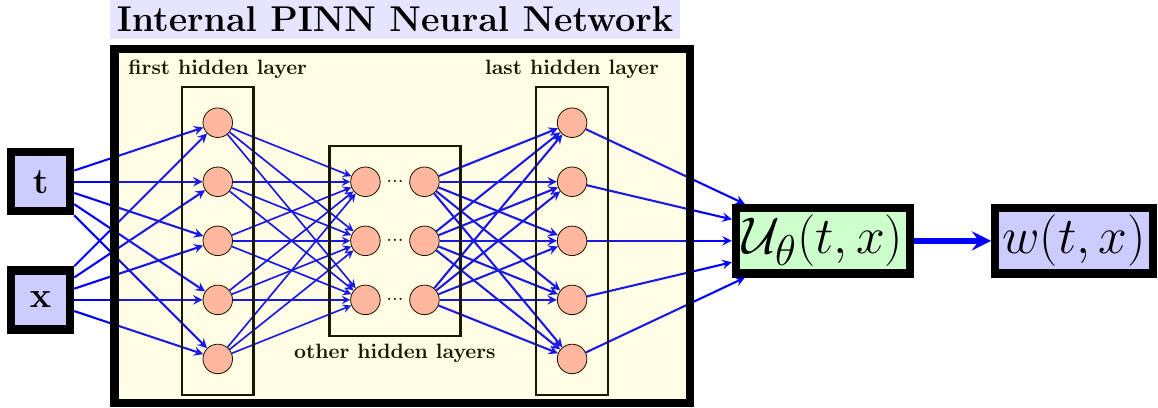}
\caption{Schematic view of a PINN. The inner structure, labeled 'internal PINN neural network', up to the $\Ucal_\theta(\cdot)$ node can follow any neural network architecture chosen by the user. The output $w(\cdot)$ is the equation error $w$ from relation~\eqref{eq:error_equation}.}
\label{fig:pinn_with_error_equation}
\end{figure}

To set notations let us recall very briefly how PINNs operate. Suppose that the evolution equation is to be solved is written in the form~:
\begin{align}
    \partial_t u &= \Gcal_t u
\label{eq:exact_equation} \\
u(0,x) & =u_0(x), \ \forall x \in \Omega \\
u(t,x) &= u_{bc}(t,x), \ \forall t\in[0,T], x \in \partial \Omega.    
\end{align}
where $u(\cdot,\cdot)$ is the main unknown, $t$ is the time variable and $x$ the spatial variable, 
$\Gcal_t$ is some operator possibly involving the 
spatial derivatives of $u$ (e.g., for the heat equation $\Gcal_t = \partial_{xx}$), $u_0(\cdot)$ and $u_{bc}(\cdot,\cdot)$ are the initial and boundary conditions respectively. When the equation is set in finite dimension such as the Lorenz system in section~\ref{sec:lorenz} then $\Omega$ will be discrete and $\partial \Omega = \emptyset$. The solution $u$ will be searched in the form of a neural network (NN) taking two inputs $t$ and $x$ and outputting an approximation $\Ucal_\theta(t,x)$ of $u(t,x)$~; $\theta$ are the
NN parameters, see figure~\ref{fig:pinn_with_error_equation}.
In order to simplify the presentation we will moreover assume as in \cite{pinn_causality,turinici24_time_sampling_PINN1} and related works, that we can construct the network $\Ucal_\theta$ such that $\Ucal_\theta$ satisfies exactly the initial and boundary conditions. For instance for the initial conditions one can simply shift the output of the NN~:
$\Ucal_\theta(t,x) \mapsto 
\Ucal_\theta(t,x)-\Ucal_\theta(0,x)+u_0(x)\cdot t$
as in \cite{pinn_causality} or  
\begin{equation}
\Ucal_\theta(t,x) \mapsto 
\Ucal_\theta(t,x)-\Ucal_\theta(0,x)+u_0(x).   \label{eq:shift_for_u0} 
\end{equation}
 as in \cite{turinici24_time_sampling_PINN1}.
In order to train the NN a loss is formulated that includes as main ingredient the equation error $w(t,x)$ defined as $w(t,x) = \partial_t \Ucal_\theta(t,x) - \Gcal_t \Ucal_\theta(t,x)$ or equivalently~:
\begin{equation}
\partial_t \Ucal_\theta(t,x) = \Gcal_t \Ucal_\theta(t,x) + w(t,x), \ 
\Ucal_\theta(0,x) = u_0(x).
\label{eq:error_equation}
\end{equation}
Our use of this last equation is slightly different from the literature, see also figure~\ref{fig:pinn_with_error_equation}~: since the derivatives are here {\bf exactly} computed using the automatic differentiation capabilities of the NN, the equation 
\eqref{eq:error_equation} is also satisfied {\bf exactly}. So we can consider the error $w$ as being the main output and $\Ucal_\theta$ as the solution of 
\eqref{eq:error_equation} that is available at no additional cost.

The NN is trained to minimize a loss functional; the original proposal in~\cite{RAISSI2019686} is 
$\int_0^T \int_\Omega | w(t,x)|^2 dx dt$ but, later on, particular weighting schemes appeared that proposed a loss of the form~:
\begin{equation}
\Lcal(\theta) = \int_0^T \int_\Omega \rho(t) | w(t,x)|^2 dx dt,
\label{eq:def_loss_rho}
\end{equation}
with the weight $\rho(t) \ge 0$ to be chosen in order to speed up the convergence and improve the quality of the output. For instance \cite{pinn_causality} use 
a discrete form of 
\begin{equation}
\rho(t) = e^{-\epsilon \int_0^t \|w(t,x)\|^2_{L^2_x}}    
\label{eq:causal_weights_formula}
\end{equation}
for some $\epsilon >0 $~; such a choice will give more weight to low values of $t$. Another proposal is 
\cite{turinici24_time_sampling_PINN1} who sets  
 $\rho(t) = e^{-\epsilon t}$ for some $\epsilon \in \R$; the intuitive reason is that errors at earlier times will accumulate into larger errors in the final output so  the solution at initial times should be computed more precisely. Although these proposals are inducing better numerical properties, they remain somehow ad-hoc. We investigate below in a principled way 
 how the weights $\rho$ contribute to the quality of the result and set it accordingly.

\section{Theoretical results} \label{sec:theory}

To design time weighting schemes we need to know how the equation error $w(t,\cdot)$ at time $t$ will impact the accuracy of the final result $\Ucal_\theta(T,\cdot)$. There is no general answer to this question but we will use a close concept, namely the {\bf Lyapunov exponent} which describes  quantitatively 
the rate of separation of infinitesimally close trajectories. The Lyapunov exponent originates from the dynamic system analysis and, for some 1D ordinary differential equation $z'(t) = f(t,z(t))$ describes how two solutions, starting from $z^1(0)$ and $z^2(0)$ diverge in the limit of large times $t$; more precisely the Lyapunov exponent is the coefficient $\lambda \in\R$ such that $| z^1(t)-z^2(t)| \sim e^{\lambda t} | z^1(0)-z^2(0)|$ (for large $t$). The determination of the exact value of the Lyapunov exponent is a difficult task in dynamical systems and ever more so in evolution PDE, but we will retain this idea. This allows to state~:
\begin{proposition}
Let $u$, $\Ucal_\theta$ as in  \eqref{eq:exact_equation} and \eqref{eq:error_equation} respectively. Let $\lambda(t)$ be such that 
\begin{equation}
\frac{\langle \Gcal_t(u(t,\cdot))- \Gcal_t(\Ucal_\theta(t,\cdot)), u(t,\cdot)- \Ucal_\theta(t,\cdot) \rangle}{ \|u- \Ucal_\theta\|^2} \le  \lambda(t).  \label{eq:definition_lambda}
\end{equation}
Then
\begin{equation}
\|u(T,\cdot)- \Ucal_\theta(T,\cdot)\| \le  \int_0^T e^{ \int_t^T\lambda(s)ds} \|w(t,\cdot)\| dt. 
\label{eq:prop_estim_error}
\end{equation}
Here the norm is euclidean when $\Omega$ is finite and $L^2$ norm otherwise.
\end{proposition}
\begin{proof} Denote $e(t) = u(t,\cdot)- \Ucal_\theta(t,\cdot)$. Using \eqref{eq:exact_equation} and \eqref{eq:error_equation}  we obtain:
\begin{align}
&  \frac{1}{2}  \frac{d}{dt} \|e(t)\|^2 = \left\langle \frac{d}{dt} e(t), e(t)\right\rangle
=  \left\langle  \Gcal_t(u)- \Gcal_t(\Ucal_\theta)- w(t,\cdot), e(t)\right\rangle
\\ &
\le \lambda (t) \left\langle  e(t), e(t)\right\rangle -
\left\langle w(t,\cdot), e(t)\right\rangle
\le \lambda (t) \| e(t)\|^2 + \|w(t,\cdot)\| \cdot \| e(t)\|
\end{align}
This means that $ \frac{d}{dt} \|e(t)\| \le  \lambda (t)\|e(t)\| + \|w(t,\cdot)\|$, and, using the notation $\Lambda(t) = e^{-\int_0^t \lambda(s) ds}$
we obtain $\frac{d}{dt} \left(\Lambda(t) \| e(t) \| \right) \le \Lambda(t) \|w(t,\cdot)\|$~; hence, since $e(0) = 0$, by integration from $0$ to $T$ we get 
$\Lambda(T) \| e(T) \| \le \int_0^T \Lambda(t)  \|w(t,\cdot)\| dt$ which gives the conclusion.
\end{proof}

The result already gives qualitative insight into how the error at the final time $T$ depends on errors at intermediary times $t\le T$, which is the main idea of Lyapunov exponents. Let us take the simple case $\Gcal_t(u) = \lambda u$ with $\lambda \in \C$ constant; in this case 
\eqref{eq:prop_estim_error} is in fact an equality.
As in \cite{pinn_causality} we note that for a chaotic or divergent system, which is characterized by a strictly positive Lyapunov exponent i.e., $Re(\lambda)>0$ the trajectories diverge exponentially with respect to error in the equation \eqref{eq:error_equation} or in the data. In this case the precision has to be large for $t$ near zero because the factor $e^{\lambda (T-t)}$ will multiply it resulting in large error in the final state. On the contrary, for values of $t$ close to $T$ the factor $e^{\lambda (T-t)}$ is relatively small and a larger error in the equation can be tolerated because it will be multiplied by a smaller factor. What is interesting to note is that this also works the other way around for $Re(\lambda)<0$ which characterize  systems converging to stable equilibria. In this case initial errors in the equation will be "washed away" by the stable dynamics and are less important and estimation \eqref{eq:error_equation} says that we can manage errors increasing {\bf exponentially} with $T-t$ as long as the exponent is smaller than $|Re(\lambda)|$ !
Finally, for the (quasi-) periodic dynamics $|\lambda|=1$, all times $t\le T$ contribute equally to the final error and no special weighting should be enforced. 

To add more actionable quantitative details to the discussion above we 
will see now how this translates into weighting schemes for $\rho(t)$; changing the weight $\rho(t)$ indicates to the training algorithm that some values of $t$ should be given priority when minimizing the error loss $w(t,\cdot)$. Of course, more computational resources are available better the algorithm will converge, so we will make our reasoning assuming a given computational level of error i.e., will assume that we can employ numerical algorithms that  make $\int_0^T \rho(t) \| w(t,\cdot)\|^2 dt$ small enough (but not zero), say equal to $C_p$ and inquire which is the best weighting scheme for this class of errors. The idea is to see how well the choice of $\rho$ {\bf guarantees} a good behavior of the final error at time $T$. Note that the training will only minimize the loss without regards as to which of the possible functions $w$ are selected, as long as 
 $\int_0^T \rho(t) \| w(t,\cdot)\|^2 dt$ reaches the desired level $C_p$. So here we will assume worse case scenario and see what is the $\rho$
 whose worse case scenario is the best. This will be formalized as looking for the $\rho$ such that 
 $\left\{ \max_{\int_0^T \rho(t) \| w(t,\cdot)\|^2 dt \le C_p} \int_0^T e^{\int_t^T \lambda(s)ds } \| w(t,\cdot)\| dt \right\}$
is smaller, see proposition below.
To gain in generality we will accept weighting schemes with irregular densities and write for a law $\eta$ on $[0,T]$ with density $\rho$ (maybe not a proper function) $\E_{\tau \sim \eta} \| w(\tau,\cdot)\|^2 dt$ instead of 
 $\int_0^T \rho(t) \| w(t,\cdot)\|^2 dt$.
 For technical reasons we will need the law $\eta$ to have finite second order moment which we write $\eta \in \Pcal^2([0,T])$ and we assume $w$ is bounded.
The following result gives precise information on the best weighting to be used~:
\begin{proposition}
Let $\lambda(s):[0,T] \to \R $ with 
$\int_0^T e^{\int_0^t \lambda(s)ds } dt < \infty$ and $w$ such that 
$t\mapsto \| w(t,\cdot)\|$ is bounded. 
Then the  problem
\begin{equation}
\min_{\eta \in \Pcal^2([0,T])}  
\left\{ \max_{w; \E_{\tau \sim \eta }[\| w(\tau,\cdot)\|^2 ] \le C_p} \int_0^T e^{\int_t^T \lambda(s)ds } \| w(t,\cdot)\| dt \right\}
\end{equation}
admits an optimum which is attained in some $\eta^{opt}$. Moreover the minimum $\eta^{opt}$ has a density $\rho^{opt}(t)$ which is proportional to 
$e^{- \int_0^t \lambda(s) ds}$ that is~:
\begin{equation}
    \rho^{opt}(t) = \frac{e^{- \int_0^t \lambda(s) ds}}{\int_0^T e^{- \int_0^t \lambda(s) ds} dt}.
\end{equation}
\end{proposition}
\begin{proof}
We first suppose that $\rho$ is regular enough and look for the solution.
Denote $\Lambda(t) = e^{- \int_0^t \lambda(s) ds}$ and write 
$ \int_0^T e^{\int_t^T \lambda(s)ds } \| w(t,\cdot)\| dt  = 
\frac{1}{\Lambda(T)} \int_0^T \Lambda(t) \| w(t,\cdot)\| dt$. Dismissing the constant 
$1/\Lambda(T)$ we are thus maximizing $\int_0^T \Lambda(t) \| w(t,\cdot)\| dt$ under the constraint $\E_{\tau \sim \eta }[\| w(\tau,\cdot)\|^2 ] \le C_p$.
The Cauchy's inequality informs that 
\begin{equation}
\left( \int_0^T \Lambda(t) \| w(t,\cdot)\| dt \right)^2 \le 
 \E_{\tau \sim \eta }[\| w(\tau,\cdot)\|^2 ] 
 \E_{\tau \sim \eta }[(\Lambda^2(\tau)/ \rho^2(\tau) ) ]  
\end{equation}
with equality only when $\Lambda(\tau)/ \sqrt{\rho(\tau)}$ and $\sqrt{\rho(t)}\| w(\tau,\cdot)\|$ are proportional i.e. 
$\|w(\tau,\cdot)\|$ is proportional to  $\Lambda(t)/\rho(t)$. In this case the maximal value is $ \int_0^T \Lambda^2(t) / \rho(t) dt $. 
So for a given $\rho$ we cannot guarantee a final error better than
$ \int_0^T \Lambda^2(t) / \rho(t) dt $ (up to a multiplicative constant depending on $C_p$ and $\Lambda(T)$ but not on $\rho$).
Now the choice of $\rho$ should set this integral to the smallest value possible under the constraint $\int \rho(t) dt = 1$  and $\rho \ge 0$. Use again the Cauchy inequality to see that 
\begin{equation}
\int_0^T \Lambda^2(t) / \rho(t) dt  = 
\int_0^T \rho(t) dt \int_0^T \Lambda^2(t) / \rho(t) dt \ge 
\left(\int_0^T \Lambda(t) dt\right)^2,
\end{equation}
which shows that we cannot do better than $\left(\int_0^T \Lambda(t) dt\right)^2$, 
with equality when $\Lambda^2(t) / \rho(t)$ and $\rho(t)$ are proportional i.e.
$\rho(t)$ is proportional to $\Lambda(t)$, which is the conclusion. 

For any other, possibly non smooth, choice of $\rho$ one can follow the same reasoning by regularization~: take smooth approximations of $\rho$ which, by the arguments above turn out to not be better than $\rho^{opt}$ so passing to the limit one obtains that no non-smooth $\rho$ can do better. 
\end{proof}
So the best weights $\rho(t)$ are proportional to 
$e^{- \int_0^t \lambda(s) ds}$; this quantity is calculated from the values of $\lambda(t)$. Ideally, one would like to have~:
\begin{equation}
    \lambda(t) = \frac{ \langle\Gcal_t(\Ucal_{\theta}(t,\cdot)) - \Gcal_t(u(t,\cdot)), \Ucal_{\theta}(t,\cdot)- 
    u(t,\cdot)\rangle}{ \| \Ucal_{\theta}(t,\cdot)- u(t,\cdot) \|^2}, 
    \label{eq:ideal_lambda}
\end{equation}
i.e., the $\lambda$ that is realizing equality in \eqref{eq:definition_lambda}. But the right hand side depends on the exact solution which is the unknown and has to be approximated.  
We will estimate the $\lambda(s)$ by taking simply~:
\begin{equation}
    \lambda_n(t) \sim \frac{ \langle\Gcal_t(\Ucal_{\theta_{n}}(t,\cdot))-\Gcal_t(0), \Ucal_{\theta_{n}}(t,\cdot)
    \rangle}{ \| \Ucal_{\theta_{n}}(t,\cdot)  \|^2}
    \label{eq:estimator_ls_ref_first}
\end{equation}
where $\theta_n$ is the value of the NN parameters after the $n$-th training step ($\Ucal_{\theta_{n}}$ is the solution candidate at this iteration). 
This amounts to approximating the exact solution (which is unknown) to zero. Although this approximation seems rough, it the most universal to be made and in practice it gives good results because we do not want exact values for $\lambda_n(t)$ but some general information e.g., on its sign and monotonicity. Of course, if additional information on the exact solution $u(t,x)$ is available this could be incorporated into  \eqref{eq:estimator_ls_ref_first}. 
Furthermore, note  that for the simple situation when 
$\Gcal_t(u) = \Acal_t(u) + \xi_t + \alpha_t u$,
with $\Acal_t$ a anti-symmetric operator, $\xi_t$ independent of $u$ and $\alpha_t \in \R$ the choice \eqref{eq:estimator_ls_ref_first} results in $\lambda_n(t) = \alpha_t$ (for all $n$) {\bf which is exact} and optimal.

\section{Numerical experiments} \label{sec:numerical}

The code for all the numerical experiments performed in this work and presented in this section is available as Github package at~: \\  {\scriptsize \url{https://github.com/gabriel-turinici/physics_informed_neural_networks_time_sampling}}.

\subsection{Lorenz system} \label{sec:lorenz}

The Lorenz system is a set of ordinary differential equations that describes a simplified model of atmospheric convection. It exhibits chaotic behavior and has been widely studied in the field of dynamical systems; it is described by the following set of ordinary differential equations:
\begin{equation}
x'(t) = \sigma (y - x), 	\ 
y'(t) = x (\rho - z) - y, 	\ 
z'(t) = x y - \beta z,
\label{eq:lorenz_system}
\end{equation}
where $x$, $y$, and $z$ are the state variables, and $\sigma$, $\rho$, and $\beta$ are parameters.
Coherent with the literature \cite{pinn_causality} we will take final time $T=0.5$\footnote{In the literature time goes up to $T=20$ but in practice it is implemented as sequential solves on $40$ time intervals of size $0.5$.} and use the classical parameter set (initially studied by Lorenz) 
$(\sigma, \rho, \beta) = (10, 28, 8/3)$ and the initial state 
$(x, y, z)(0)=(1,1,1)$. This test case is notoriously difficult to solve with PINN  (the default scheme does not even converge, see below) because of its very high sensitivity with respect to errors in the data and equation.
\begin{figure}
\centering
\includegraphics[width=0.49\linewidth]{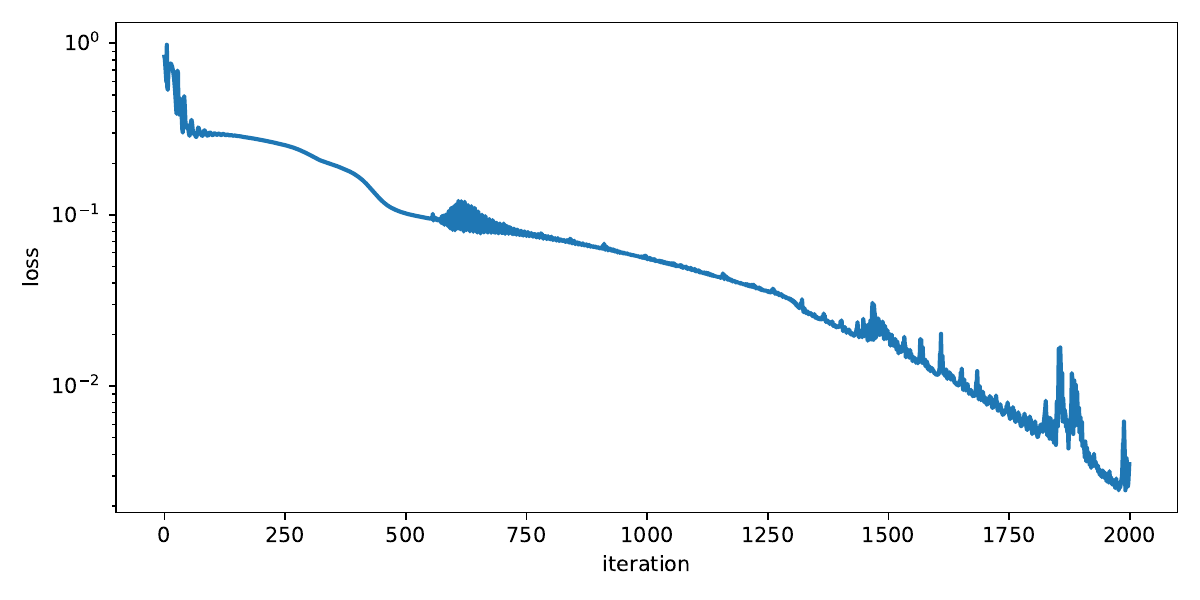}
\includegraphics[width=0.49\linewidth]{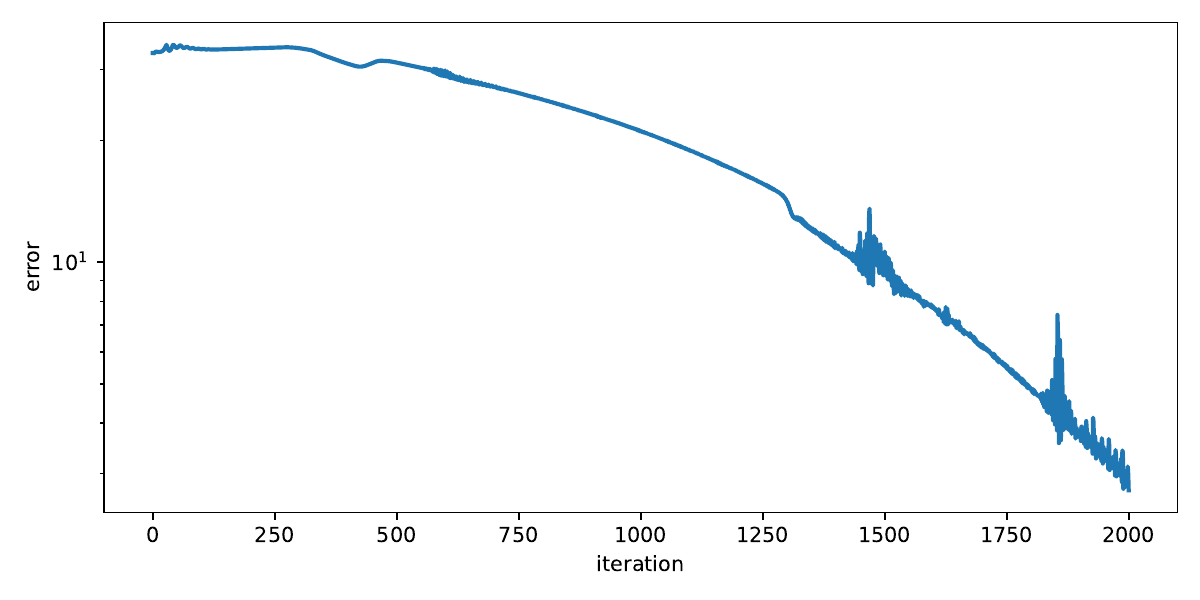}
\caption{Lorenz system, convergence of our procedure. {\bf Left} the loss convergence. {\bf Right} the convergence of the error at time $T$ (see text).}
\label{fig:pinn_lyapunov_convergence}
\end{figure}

\begin{figure}
\centering
\includegraphics[width=0.99\linewidth]{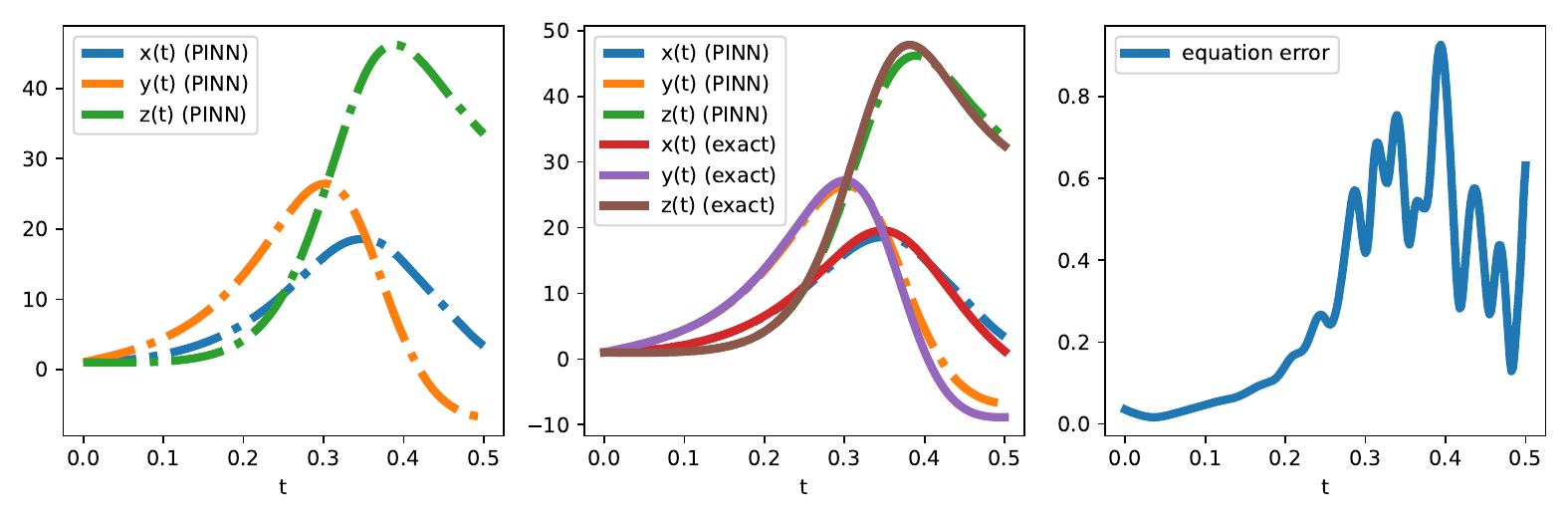}
\caption{Lorenz system. {\bf Left}: the solution given by our procedure; 
{\bf middle} the comparison between our solution (dash dotted) with reference solution (solid lines): {\bf right} the equation error $t\mapsto \|w(t,\cdot)\|$.}
\label{fig:pinn_lyapunov_solution}
\end{figure}

\begin{figure}
\centering
\includegraphics[width=0.4\linewidth]{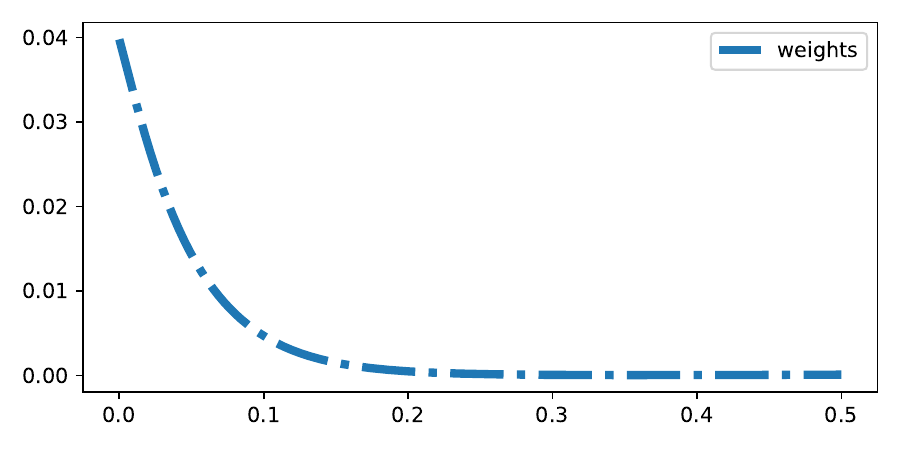}
\includegraphics[width=0.4\linewidth]{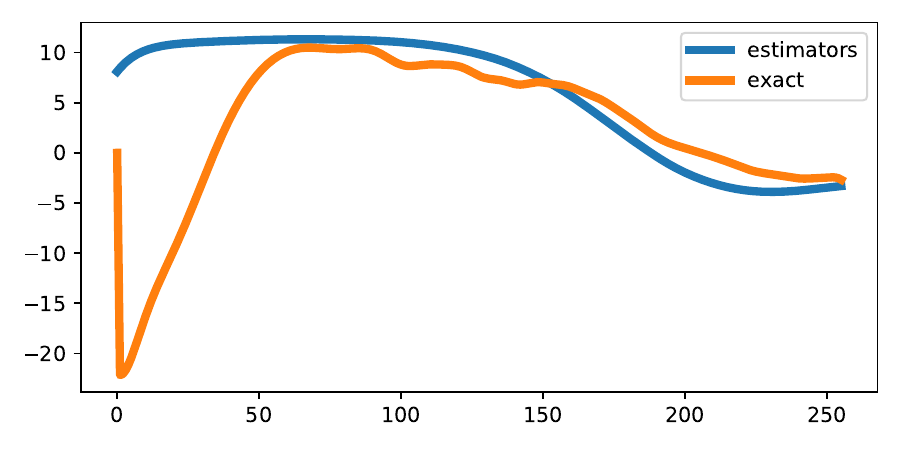}

\caption{Lorenz system. {\bf left} the final wights selected by our procedure (abscissas are times $t$, ordinates the weights); {\bf right} the Lyapunov estimator in~\eqref{eq:estimator_ls_ref_first} 
versus exact values in equation~\eqref{eq:ideal_lambda} (abscissas are indices of the time grid).}
\label{fig:pinn_lyapunov_weights}
\end{figure}

\begin{figure}
\centering
\includegraphics[width=0.49\linewidth]{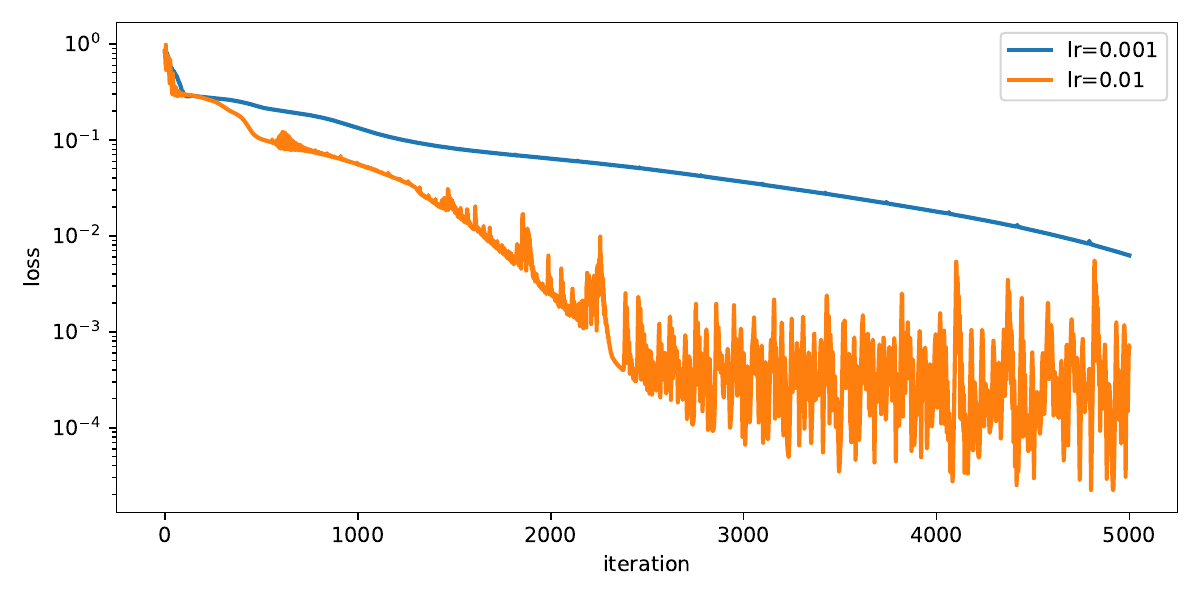}
\includegraphics[width=0.49\linewidth]{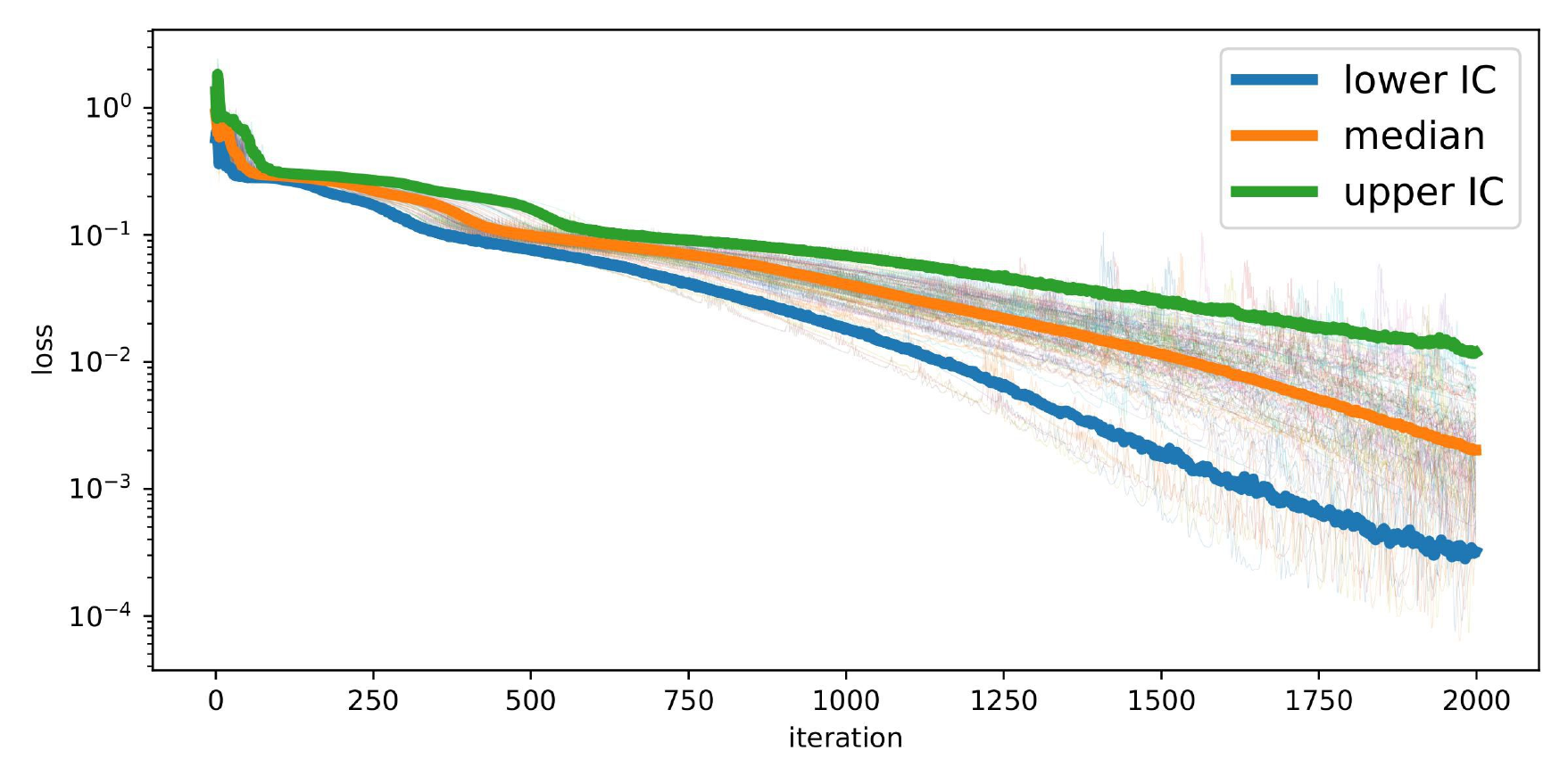}
\caption{Lorenz system, {\bf Left~:} the convergence of the loss for different learning rates $0.001$ (default) and $0.01$. As expected the value $0.001$ gives slower convergence but smaller final variance (oscillations) at the solution;
{\bf right~:} the median and $95\%$ confidence interval of the loss 
(thicker lines) together with the first $100$ individual trajectories (finer lines).}
\label{fig:pinn_lyapunov_compare_lr_loss}
\end{figure}

The reference solution, considered exact, is the one obtained using the Scipy package {\scriptsize\verb|scipy.integrate.odeint|} routine with default settings. 
For the initial condition we used the shift  in~\eqref{eq:shift_for_u0}. 
The NN has $5$ hidden fully connected layers of $20$ neurons each ('tanh' activation) and a final FC layer with output dimension $3$ (to match the unknown dimension) with no activation. All layer initializations are Glorot Uniform. The NN has $1'783$ trainable parameters (which compares very favorably with $1'054'723$ training parameters used by~\cite{pinn_causality} who employ $5$ hidden FC layers with $512$ neurons). We used a uniform time grid of $256$ points (as in \cite{pinn_causality}) and $2000$ iterations of the Adam optimizer; initially we tested with the default learning rate ($0.001$) but to accelerate convergence we then chose everywhere a learning rate of $0.01$ (only exception being  figure~\ref{fig:pinn_lyapunov_compare_lr_loss}  where we compare these two learning rates). All other optimizer parameters are the TensorFlow version 2.15 defaults. A run of the procedure takes 80 seconds on a T4 GPU.

We first show convergence results in figure~\ref{fig:pinn_lyapunov_convergence}. It is seen that the loss reaches low values (and will improve if more iterations are allowed); at the same time we plot (right figure) also the error at final time $T$ between the reference solution and the PINN solution, that also behave well and shows that convergence occurred. This is to be compared with figure~\ref{fig:pinn_lyapunov_solution} where the numerical solution shows high quality match with the baseline solution. We also checked (not shown here) that given more iteration the quality improves further (but then we cannot distinguish graphically the two).

We now pass to the inner workings of our procedure. First in figure~\ref{fig:pinn_lyapunov_solution} (right) we note that the equation error is indeed of intuitive form: since the system is very sensitive to errors, it is best to set error to be lower for small $t$ and larger at final times. 
This is coherent with the weights selected (automatically) by our algorithm, depicted in figure \ref{fig:pinn_lyapunov_weights} left, that are indeed decreasing. The weight assignment is the result of the estimators \eqref{eq:estimator_ls_ref_first} plotted in figure \ref{fig:pinn_lyapunov_weights} right. We plot both our estimator and the ground truth \eqref{eq:ideal_lambda} (that requires knowledge of the exact solution). The agreement is remarkable and it is seen that our estimator \eqref{eq:estimator_ls_ref_first} {\bf captures the good order of magnitude and sign} while retaining a more smooth behavior.

As discussed above, we also investigated the influence of the learning rates; we plot in figure \ref{fig:pinn_lyapunov_compare_lr_loss} left 
a comparison of the losses during the optimization with two different learning rates, $0.001$ (TensorFlow Adam optimized default) and $0.01$. As expected from the general theory of stochastic optimization it is seen that the larger learning rate gives faster convergence but also larger final variance (oscillations) at the solution. 

Finally, in figure~\ref{fig:pinn_lyapunov_compare_lr_loss} right we plot a confidence interval for the loss (learning rate set to $0.01$ for all) after $1000$ independent runs of the procedure. The variability is due to the initialization of the NN layers and is normal to have runs taking longer time. For all quantiles the decrease in loss is systematic. 

\begin{figure}
\centering
\includegraphics[width=0.99\linewidth]{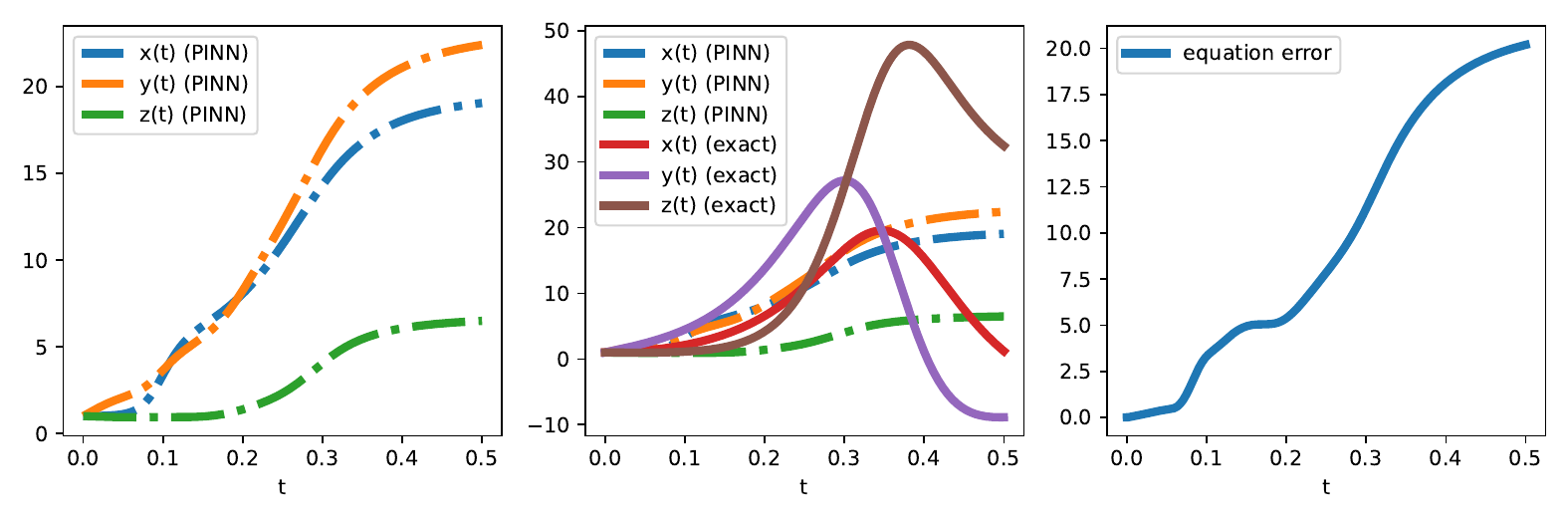}
\caption{Lorenz system. {\bf Left}: the solution given by the causal procedure in \cite{pinn_causality}; 
{\bf middle} the comparison between the solution (dash dotted) with reference solution (solid lines): {\bf right} the equation error.}
\label{fig:pinn_causal_solution}
\end{figure}

\begin{figure}
\centering
\includegraphics[width=0.49\linewidth]{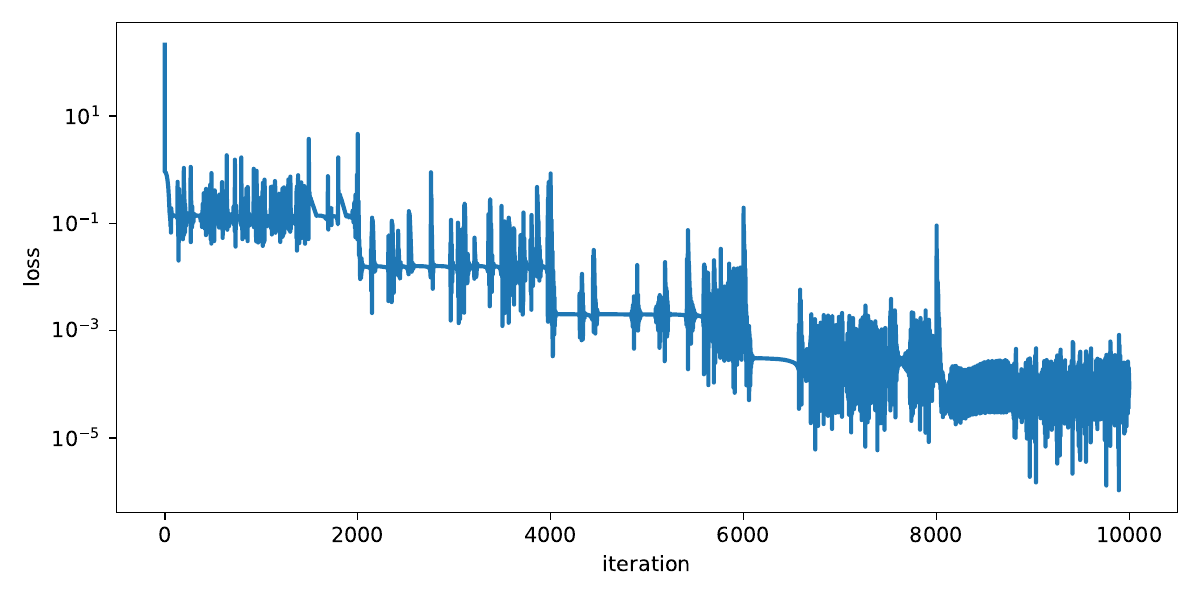}
\includegraphics[width=0.49\linewidth]{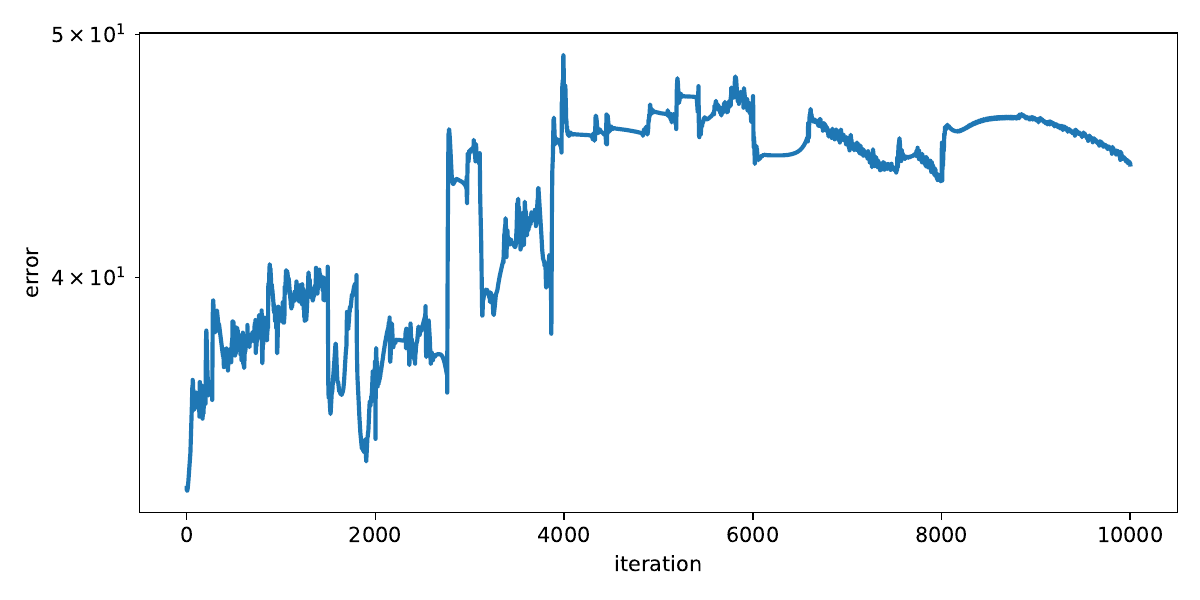}
\caption{Lorenz system, convergence of the 'causal' procedure. {\bf Left} the loss convergence. {\bf Right} the evolution of the error at time $T$.}
\label{fig:pinn_causal_convergence}
\end{figure}

We now turn to comparing these results with equivalent results for two other time weighting algorithms, \cite{pinn_causality} (cumulative exponential error loss \eqref{eq:causal_weights_formula}) and \cite{turinici24_time_sampling_PINN1} (exponential weighting $\rho(t) \sim e^{-\lambda t}$). 
We do not even compare with uniform time-weighting (which is the default PINN protocol) because this is known to not work well on Lorenz (and most chaotic) systems. The interested reader can nevertheless find in the supplementary material section~\ref{sec:lorenz_notokstandardPINN} this result.

For both we used exactly the same NN and optimizer settings.
We seeded the random number generation of numpy and TensorFlow with the same seed as in the previous case so all procedures start from the same initial network weights.
The 'causal' procedure seem to not have converged yet cf. figure~\ref{fig:pinn_causal_solution} even if we gave it $5$ times more iterations. We did so because this procedure needs to change the value of $\epsilon$ in a schedule involving $5$ preset levels in the list $0.01,0.1, 1, 10, 100$. To each level we gave $2000$ iterations (which is the total number of iterations used by our procedure). We checked that if given even more time the protocol does indeed converges. Obviously it attempts to exploit the same idea of using larger weights for $t$ small resulting in lower equations errors at $t$ small and larger errors for $t$ large (cf. figure~\ref{fig:pinn_causal_solution} rightmost plot).
In figure~\ref{fig:pinn_causal_convergence} left we plot the convergence of the procedure (the $5$ $\epsilon$ levels are visible on left plot). Note that, since the Lorenz system is close to chaotic, significant equation errors of the early stages of the evolution equation destroy completely any confidence in the solution for final time $T$ as is shown in figure~\ref{fig:pinn_causal_convergence} right, where the final error between the exact and numerical solution does not appear to improve much. The loss is indeed decreasing but it does not convey enough relevant information to orient the convergence. See supplementary material for further explanation of the weights behavior in this procedure, which appear sensitive to the precise number of iterations run for each  value of $\epsilon$.

\begin{figure}
\centering
\includegraphics[width=0.99\linewidth]{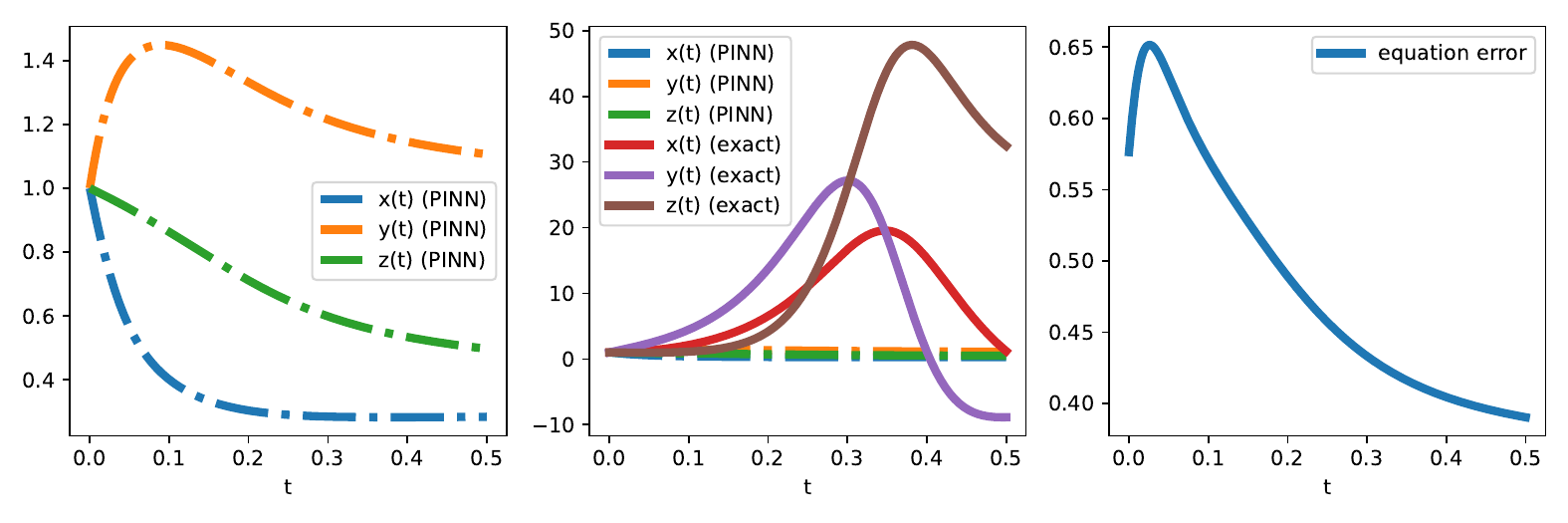}
\includegraphics[width=0.99\linewidth]{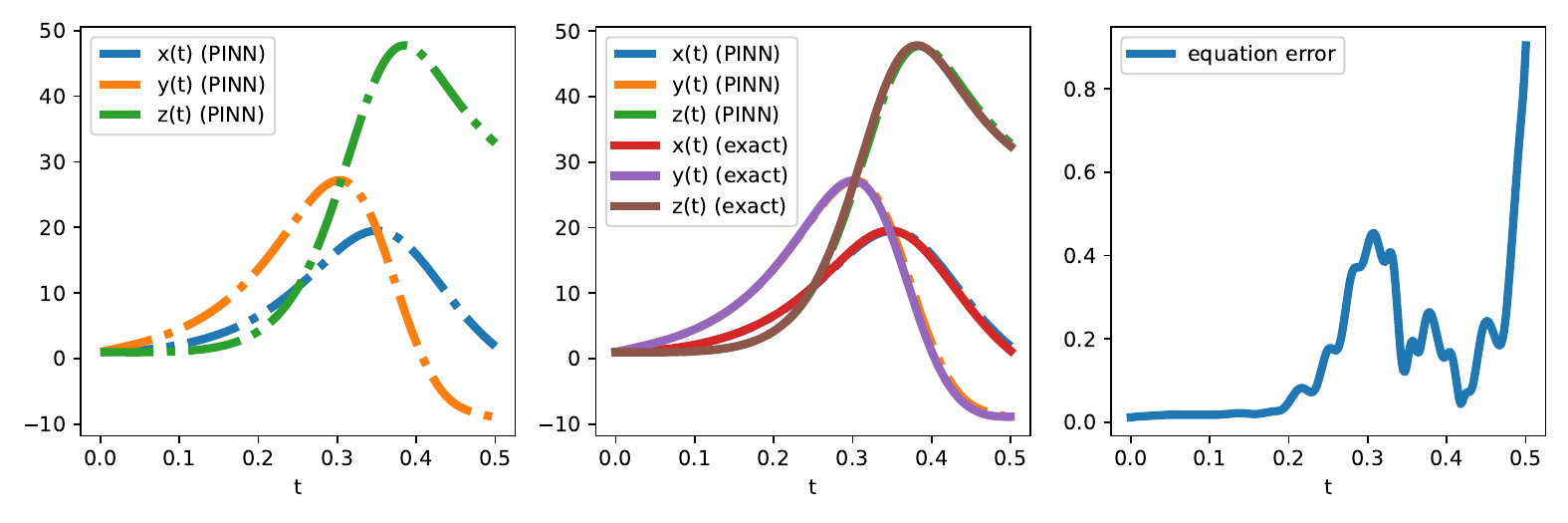}
\caption{Lorenz system; the solution given by the exponential weighting procedure in \cite{turinici24_time_sampling_PINN1} for $\lambda=10.0$ (first row) and $\lambda=13.8$ (second row). 
{\bf left column}: the numerical solution
{\bf middle column} the comparison between the solution (dash dotted) with reference solution (solid lines): {\bf right column} the equation error.}
\label{fig:pinn_cst10_solution}
\end{figure}

We compare now with the exponential time weighting scheme in ~\cite{turinici24_time_sampling_PINN1}. The procedure has a hyper-parameter $\lambda$  that give the exponential decay of the weights. We tested several choices of parameters 
but will only show results for two of them,
$\lambda=10$ (which has the good order of magnitude) and 
$\lambda=13.8$; this latter value was chosen to have a posteriori the same average decay as our procedure (and thus should give comparable results). The results in figure ~\ref{fig:pinn_cst10_solution} for $\lambda=10$ show that the procedure is sensitive to the choice of the $\lambda$ parameter; when this parameter is given by some 'oracle' then the solution is much improved. When $\lambda$ is farther away from the optimal value the results degrade.

\subsection{The Burgers' equation} \label{sec:burgers}
The Burgers' equation models the one-dimensional flow of viscous  fluid and reads~:
\begin{align}
& \frac{\partial u}{\partial t} + u \frac{\partial u}{\partial x} = \nu \frac{\partial^2 u}{\partial x^2}, \ t\in ]0,T], \ x\in]-1,1[ \label{eq:burgers} \\
& u(0,x) = -\sin(\pi x) \label{eq:burgers_ic} \\
& u(t,\pm 1)=0, \ \forall t \in [0,T]. \label{eq:burgers_bc}
\end{align}
where $u(x, t)$ is the fluid velocity, $\nu=0.01/\pi$ is a real positive constant called the viscosity coefficient, $x$ and $t$ are spatial and temporal variables respectively. Total time is $T=1.0$.
These are supplemented by the initial 
\eqref{eq:burgers_ic} and boundary \eqref{eq:burgers_bc} conditions. For the initial condition we used the shift described in  \eqref{eq:shift_for_u0}; on the other hand we did nothing special for the boundary conditions which were implemented by adding a corresponding term in the loss as it is classic in these cases \cite{RAISSI2019686}.

For the 'exact' baseline, as this problem is known to be difficult to solve, a first proposal was obtained with a  
finite difference scheme combining two half-step  Crank-Nicholson propagators for the viscosity term  with a Lax-Friedrichs  propagator for the transport part. A different computation can be proposed through the exact formula (2.5) used by \cite{burgers_spectral_analytical_1986} combined with Hermite polynomial quadrature ; see the supplementary material  \ref{annex:fd_for_burgers} for all scheme details and analytic formula. We checked that both agree up to $1.0e-4$ and used quadrature analytic formula if precision beyond this was required.

As in \cite{RAISSI2019686} and \cite{self_adaptive_pinn2023} 
the NN architecture uses $8$ fully connected hidden layers of $20$ neurons each ($3021$ parameters to train) and learning rate of $0.01$; we define a uniform grid of $50$ points in time and $25$ in space (which gives a total number of interior collocation points of $1250$, almost an order of magnitude less than in the references cited). Procedure is run for $3000$ iterations in $387$ seconds on a T4 GPU.

\begin{figure}
\centering
\includegraphics[width=0.99\linewidth]{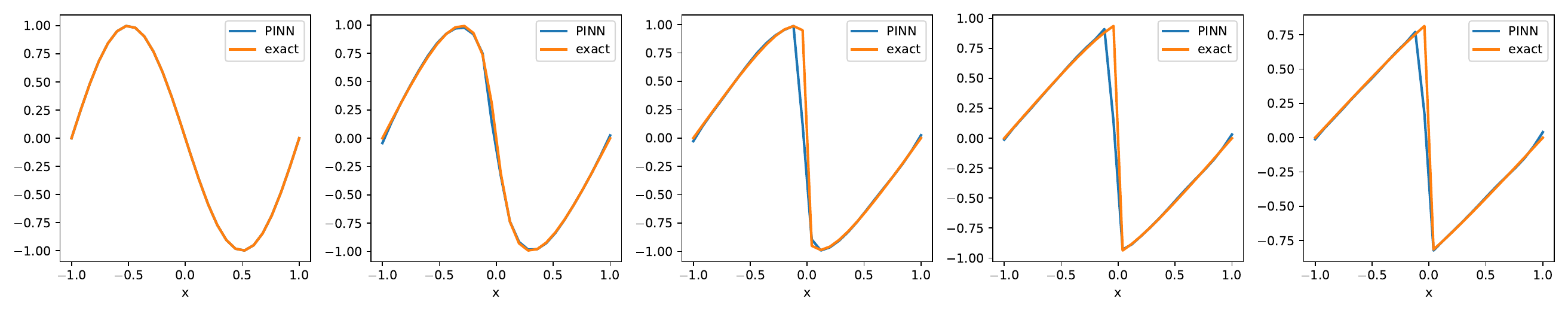}
\includegraphics[width=0.99\linewidth]{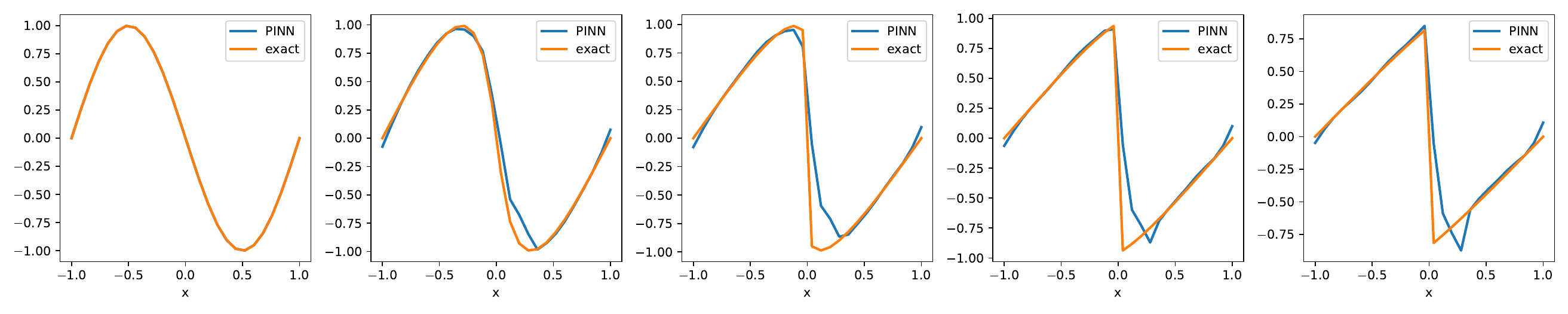}
\includegraphics[width=0.99\linewidth]{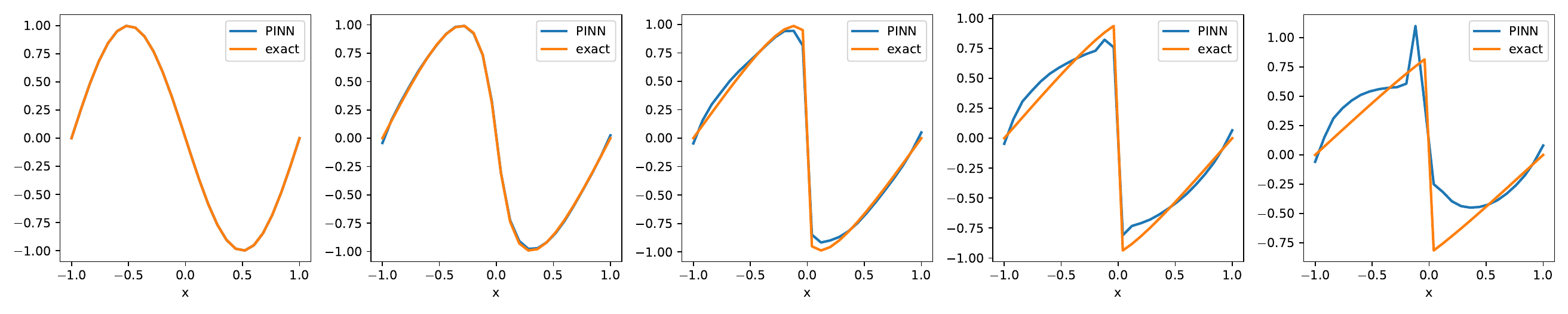}

\caption{Comparison between the exact and numerical solutions of the Burgers' equation; 
first row : solution given by our procedure; second row: solution for standard (uniform) time weighting; third row: solution for the causal \cite{pinn_causality} procedure.
Each column is another time step.}
\label{fig:pinn_burgers_solutions}
\end{figure}

We plot in figure~\ref{fig:pinn_burgers_solutions} the results for our procedure, the causal procedure \cite{pinn_causality} and the standard PINN \cite{RAISSI2019686} approach. Compared with the Lorenz system this is a non-chaotic case and all algorithms will eventually converge and give good results. The plot confirms that our procedure has some merits in using insights from the system during the initial iterations and improve efficacy. Data in supplementary material shows that this is realized by some counter-intuitive behavior of {\bf over-weighting} final instants when $t$ is close to $T$ because at this point that the discontinuity is created (and detected by our procedure).

\section{Limitations} \label{sec:limitations}
Although the theoretical and empirical results appear encouraging, further work could improve some aspects of the procedure. The most apparent is the estimation of the Lyapunov exponents, which is now done in a very crude way; this could be refined or, if not possible, at least construct some trust regions to inform when these estimations are reliable or not.

Another aspect is  the interaction between the weights dynamic and the solution. This is common to all adaptive weights procedures but it may concern some cases where the weights given by \eqref{eq:estimator_ls_ref_first} induce instabilities in the solution preventing convergence (in the same vein as the well-known GAN "mode collapse" problem). An analysis of whether this can happen or how to prevent it could be required.

Finally, for the problems where the canonical, simple implementation of the PINN procedure already works well this algorithm may cause slight numerical computational overhead.

\section{Conclusion} \label{sec:conclusion}

In this work, we presented a new approach to treat the time dimension within the framework of Physics-Informed Neural Network. We started from the observation that time instants are  not permutable and depending on the regime (chaotic, periodic, or stably convergent) errors in earlier instants affect differently the quality of the outcome. We proposed a theoretical explanation based on Lyapunov exponents and then a new algorithm built from this insight.

The procedure was then tested on two different cases, one chaotic (Lorenz system) and one stable but that induces singularities (Burgers' equation), both known to require careful numerical treatment. The proposed algorithm not only showcases robustness and practical efficiency but also addresses some limitations of earlier works that lacked a principled and theoretically grounded foundation.

While the method shows promise, several areas need improvement or future work. Notably, the estimation of Lyapunov exponents is currently crude and could benefit from refinement or the addition of trust regions. Additionally, the dynamic interaction between adaptive weights and the solution might cause instabilities, similar to GAN mode collapse, and warrants further analysis. Lastly, for problems where standard PINN methods work well,
 this approach may introduce a slight computational overhead, which seems justified by the added robustness with respect to the dynamical regimes that it can treat.
 
Of course, while our procedure performs well across various benchmarks, it is designed as a complementary addition to the existing methods rather than a one-size-fits-all solution. We believe this work contributes meaningfully to the field, offering a solid foundation for future research and application development.

\newpage 

\appendix

\section{Supplementary material}

\subsection{Standard PINN (no time weighting) results for the Lorenz system} \label{sec:lorenz_notokstandardPINN}

\begin{figure}
\centering
\includegraphics[width=0.99\linewidth]{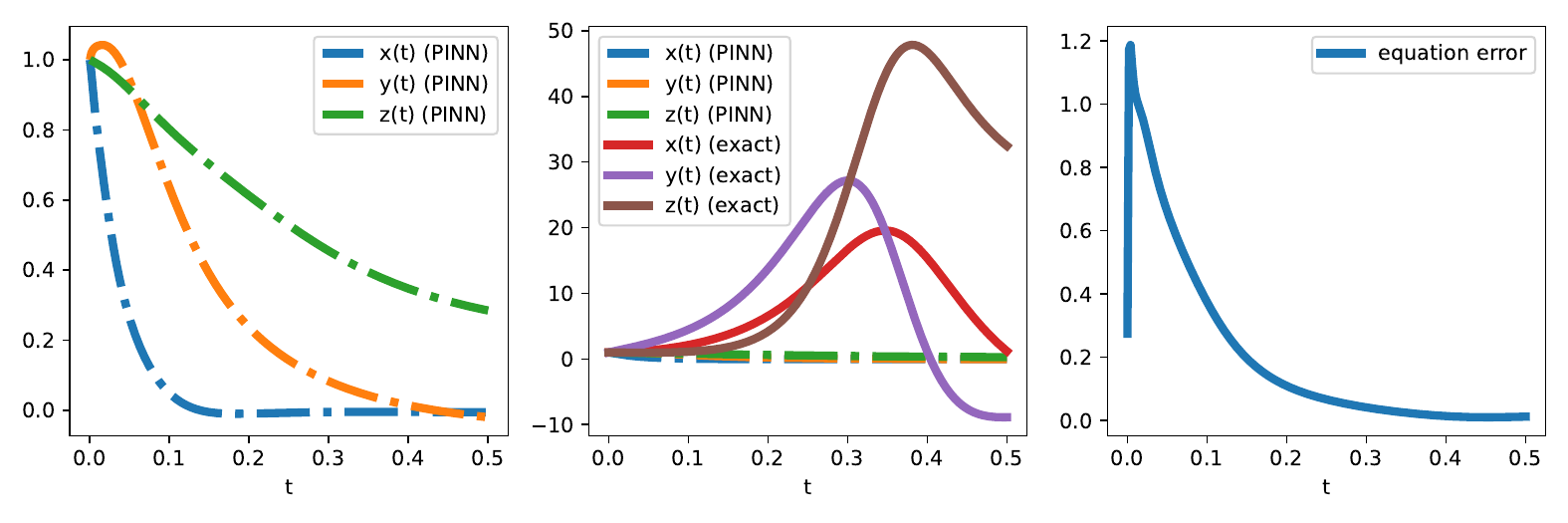}
\includegraphics[width=0.5\linewidth]{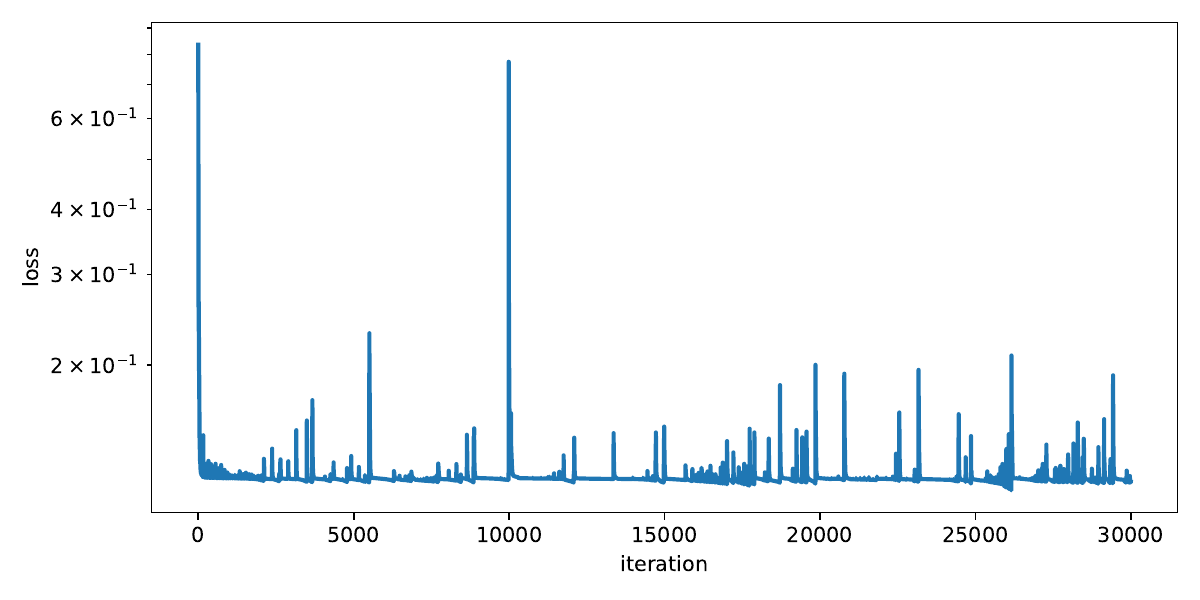}
\caption{Lorenz system. {\bf Top left}: the solution given by the canonical procedure (uniform time weighting); 
{\bf top middle} the comparison between the solution (dash dotted) and the reference solution (solid lines): {\bf top right} the equation error.
{\bf Bottom~:} the loss evolution.
}
\label{fig:pinn_cst_lambda0_10kiter}
\end{figure}

As announced in section~\ref{sec:lorenz} we plot in figure \ref{fig:pinn_cst_lambda0_10kiter} the solution of the procedure for the situation when the weights are not optimized at all but left all uniform. This case was allowed $30'000$ iterations ($10$ times more than our standard setting) but did not even start to converge.

\subsection{Exact solution for Burgers' equation}\label{annex:fd_for_burgers}
To compare the PINN solution with a reference ground truth we solved Burger's equation with a finite difference scheme and compared with an analytic formula with Hermite-Gauss integration.

\subsubsection{Finite differences scheme for Burgers' equation}\label{annex:fd_for_burgersFD}
The finite difference scheme works on a spatial domain
discretized with $nx = 400+1$ spatial grid points (counting both segment extremities $\pm 1$), 
i.e. $\Delta x =2/400$ 
and $nt = 700$ time steps ($\Delta t = 1/700$);  each time step is implementing the following split-operator technique~:

- first a Crank-Nicholson (CN) scheme over a $\Delta t/2$ time step (taking into account the boundary conditions) for the diffusion part i.e. the solution by CN of the equation $\partial_t u = \nu \partial_{xx}u$ 

- then a $\Delta t$ step of a Lax-Friedrichs scheme for the viscosity term, i.e. for the equation
$\partial_t u + u \partial_x u = 0$.

- then a final Crank-Nicholson (CN) scheme over a $\Delta t/2$ time step (taking into account the boundary conditions) for the diffusion part

The splitting technique is known to be of high order in time \cite{strang_split_operator,hubbard_splitting_method78,yanenko_splittin_1971} 
which leaves the Crank-Nicholson part (second order in time) and Lax-Friedrichs scheme (first order in time and  space) as limiting factors. As for stability it is known that CN is unconditionally stable for the heat diffusion and since the matrix for the Lax-Friedrichs scheme is tridiagonal by Gersgorin theorem it is stable when $ \max(|u|) \cdot \Delta t/ \Delta x<1$.

In practice, if $V\in \R^{nx+1}$  is the current solution with $V_j$ the component  representing the solution at point $x=-1 + j\cdot \Delta x)$, the CN propagation for a $\Delta t/2$ time step means solving the linear system in $V^{next}$~: 
\begin{eqnarray}
& \ &     \frac{V^{next}_j- V^{next}_j}{\Delta t/2} = \frac{\nu}{2} \left(
     \frac{V_{j+1}+ V_{j-1}-2 V_j}{\Delta x^2}  +
    \frac{V^{next}_{j+1}+ V^{next}_{j-1}-2 V^{next}_j}{\Delta x^2}  \right)
    \nonumber \\ & \ &    V^{next}_0=0=V^{next}_{nx}.
\end{eqnarray}

The  Lax-Friedrichs scheme means replacing $V$ with~:
\begin{equation}
    \frac{V^{next}_j- \frac{V_{j+1}+ V_{j-1}}{2}}{\Delta t} +  \left(
     \frac{ \frac{V_{j+1}^2}{2}- \frac{V_{j-1}^2}{2}}{2 \Delta x}  \right)=0, 
\ V^{next}_0=0=V^{next}_{nx}.
\end{equation}
Once this is solved the resulting 2D solution matrix is used as input to a 2D interpolation routine to obtain a linear interpolation function that is able to output values for any other point not necessarily on the space-time grid used by the finite difference resolution.

\subsubsection{Analytic formula for the solution of the Burgers' equation}\label{annex:fd_for_burgers_analytic}

Another way to find a solution is to used the analytic formula (2.5) from \cite{burgers_spectral_analytical_1986}. It reads
\begin{equation}
    u(t,x) = \frac{\int_\R u_0(x-\eta) \zeta(x-\eta) e^{-\eta^2/(4\nu t)} d \eta }{\int_\R \zeta(x-\eta) e^{-\eta^2/(4\nu t)} d \eta }, \textrm{ where } \zeta(y) = e^{-\cos(\pi y) / (2 \pi \nu)}.  
\end{equation}
The integrals can be computed using Hermite-Gauss quadrature and in practice we used $50$-th order formulas after a change of variables $\eta = \eta' \sqrt{4 \nu t }$. For $t < 10^{-10}$ we used a first order Taylor development with respect to $t$ using the exact formula for the time derivative from the Burgers' equation.

\subsection{Behavior of weights of the causal procedure (Lorenz system)}
 
\begin{figure}
\centering
\includegraphics[width=0.99\linewidth]{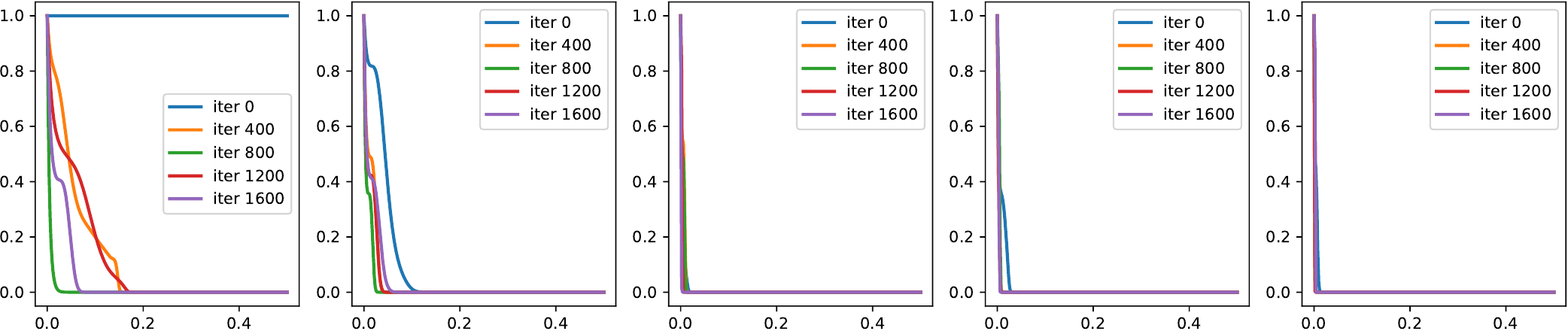}
\caption{Lorenz system, evolution  of the 
weights used by the 'causal' procedure. Each plot corresponds to a value $\epsilon$ in the list $[0.01, 0.1, 1, 10, 100]$ iterated by the procedure.}
\label{fig:pinn_causal_convergence_weights_history}
\end{figure}

To explore the convergence properties of the causal protocol \cite{pinn_causality} we looked at the history of the weights used by the procedure during the resolution of the Lorenz system, see comments in section~\ref{sec:lorenz}; these are plotted in figure~\ref{fig:pinn_causal_convergence_weights_history}.
The algorithm's protocol iterates the $\epsilon$ parameter ; at each change the weights associated with latter times are smaller which causes deterioration of the quality there. So the algorithm can only advance to the next value of $\epsilon$ only after the equation is satisfactory solved at initial times.

\subsection{Weights for Burgers' equation}
\begin{figure}
\centering
\includegraphics[width=0.32\linewidth]{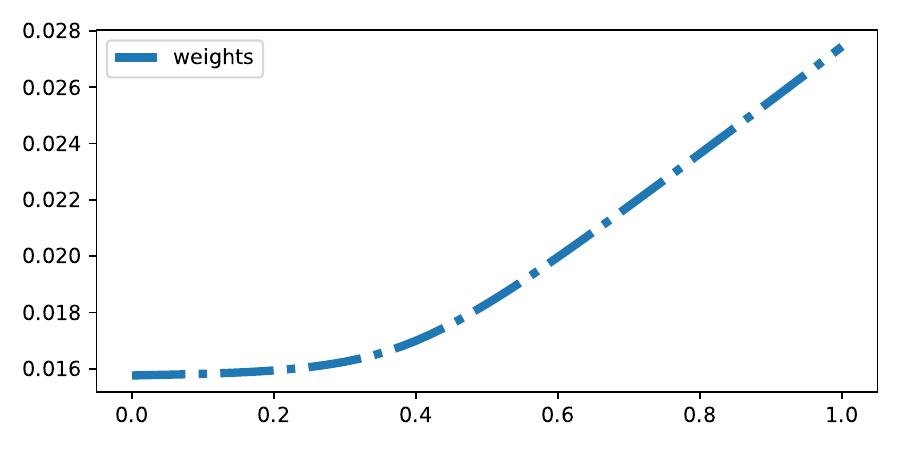}
\includegraphics[width=0.32\linewidth]{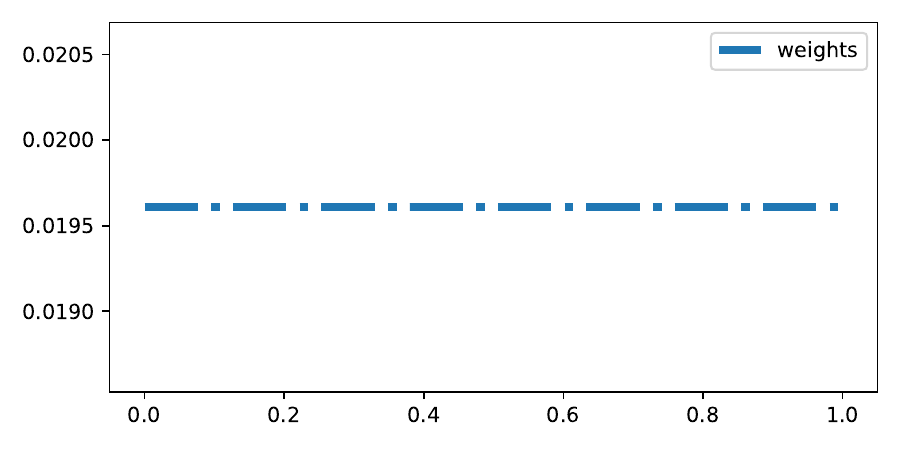}
\includegraphics[width=0.32\linewidth]{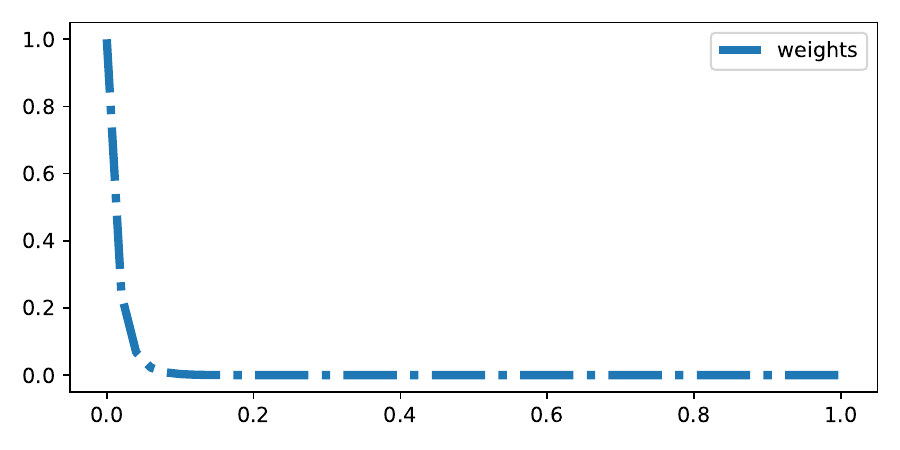}

\caption{The weights of several algorithms for the Burgers' equation; 
first row : our procedure; second row: standard (uniform) time weighting; third row: the causal \cite{pinn_causality} procedure. Compare with errors in the solution in figure~\ref{fig:pinn_burgers_solutions}.}
\label{fig:pinn_burgers_weights}
\end{figure}

We compare in figure~\ref{fig:pinn_burgers_weights} the final weights of the three procedures considered for the Burgers' equation, we note that our proposal over-weights final times which is when the discontinuity appears. This is not the case of the other procedures~; this behavior explains the results in last columns of figure~\ref{fig:pinn_burgers_solutions} (rows $2$ and $3$) that clearly lack precision at that point in space.

\end{document}